\documentclass[letterpaper, 10 pt, conference]{ieeeconf}  

\IEEEoverridecommandlockouts                              

\usepackage[utf8]{inputenc}
\usepackage[T1]{fontenc}
\usepackage[hidelinks]{hyperref}
\usepackage[style=ieee,hyperref,natbib=true,backend=bibtex,firstinits,doi=false,%
     mincitenames=1,maxcitenames=2,maxbibnames=99,sorting=none,terseinits=false,hyperref=true]{biblatex}
\bibliography{references.bib}
\renewbibmacro*{bbx:savehash}{}
\defbibheading{bibliography}[\bibname]{\section*{References}}

\usepackage{url}
\usepackage{graphicx}
\usepackage[usenames,dvipsnames]{xcolor}
\usepackage[draft,inline,nomargin]{fixme}
\fxsetup{theme=color}

\usepackage{amsmath}
\usepackage{amssymb}
\usepackage{bbm}
\usepackage{textcomp}
\usepackage{siunitx}

\usepackage{booktabs}
\usepackage{threeparttable}
\usepackage{multirow}

\usepackage[capitalize]{cleveref}

\usepackage{tikz}
\usepackage{tikz-3dplot}
\usepackage[normalem]{ulem}   
\usetikzlibrary{arrows}
\usetikzlibrary{positioning,calc}
\usetikzlibrary{decorations.pathreplacing}
\usetikzlibrary{decorations.markings}
\usetikzlibrary{fit}
\usetikzlibrary{shapes.callouts}
\usetikzlibrary{shapes.geometric}
\usetikzlibrary{matrix}
\usetikzlibrary{spy}
\usepackage{setspace}

\usepackage{pgfplots}
\pgfplotsset{compat=1.9}
\usepgfplotslibrary{groupplots}
\usepgfplotslibrary{units}
\usepgfplotslibrary{colorbrewer}
\usepackage{pgfplotstable}

\usepackage{balance}

\usepackage{blindtext}

\usepackage{ifthen}
\usepackage{tabularx}
\usepackage{makecell}

\usepackage{xcolor}
\usepackage{fixme}
\fxsetup{
    status=draft,
    author=,
    layout=inline,
    theme=color
}

\usepackage[export]{adjustbox}

\IfFileExists{graphicscache.sty}{\usepackage{graphicscache}}

\usepackage{placeins}
\usepackage{dblfloatfix}
\newcommand{\rotateRPY}[3]
{   \pgfmathsetmacro{\rollangle}{#1}
    \pgfmathsetmacro{\pitchangle}{#2}
    \pgfmathsetmacro{\yawangle}{#3}

    \pgfmathsetmacro{\newxx}{cos(\yawangle)*cos(\pitchangle)}
    \pgfmathsetmacro{\newxy}{sin(\yawangle)*cos(\pitchangle)}
    \pgfmathsetmacro{\newxz}{-sin(\pitchangle)}
    \path (\newxx,\newxy,\newxz);
    \pgfgetlastxy{\nxx}{\nxy};

    \pgfmathsetmacro{\newyx}{cos(\yawangle)*sin(\pitchangle)*sin(\rollangle)-sin(\yawangle)*cos(\rollangle)}
    \pgfmathsetmacro{\newyy}{sin(\yawangle)*sin(\pitchangle)*sin(\rollangle)+ cos(\yawangle)*cos(\rollangle)}
    \pgfmathsetmacro{\newyz}{cos(\pitchangle)*sin(\rollangle)}
    \path (\newyx,\newyy,\newyz);
    \pgfgetlastxy{\nyx}{\nyy};

    \pgfmathsetmacro{\newzx}{cos(\yawangle)*sin(\pitchangle)*cos(\rollangle)+ sin(\yawangle)*sin(\rollangle)}
    \pgfmathsetmacro{\newzy}{sin(\yawangle)*sin(\pitchangle)*cos(\rollangle)-cos(\yawangle)*sin(\rollangle)}
    \pgfmathsetmacro{\newzz}{cos(\pitchangle)*cos(\rollangle)}
    \path (\newzx,\newzy,\newzz);
    \pgfgetlastxy{\nzx}{\nzy};
}

\tikzset{RPY/.style={x={(\nxx,\nxy)},y={(\nyx,\nyy)},z={(\nzx,\nzy)}}}

\title{\LARGE \bf
MOTPose: Multi-object 6D Pose Estimation for Dynamic Video Sequences using Attention-based Temporal Fusion
}

\author{Arul Selvam Periyasamy and Sven Behnke
\thanks{All authors are with the Autonomous Intelligent Systems group, 
	Computer Science Institute VI -- Intelligent Systems and Robotics -- and the Center for Robotics and the Lamarr Institute for Machine Learning and Artificial Intelligence, University of Bonn, Germany; {\tt periyasa@ais.uni-bonn.de}}%
}

\begin{document}

\maketitle

\begin{abstract}
    Cluttered bin-picking environments are challenging for pose estimation models.
    Despite the impressive progress enabled by deep learning, single-view RGB pose estimation models perform poorly in cluttered dynamic environments.
    Imbuing the rich temporal information contained in the video of scenes has the potential to enhance models' ability to deal with the adverse effects of occlusion and the dynamic nature of the environments.
    Moreover, joint object detection and pose estimation models are better suited to leverage the co-dependent nature of the tasks for improving the accuracy of both tasks.
    To this end, we propose attention-based temporal fusion for multi-object 6D pose estimation that accumulates information across multiple frames of a video sequence.
    Our MOTPose method takes a sequence of images as input and performs joint object detection and pose estimation for all objects in one forward pass. 
    It learns to aggregate both object embeddings and object parameters over multiple time steps using cross-attention-based fusion modules. 
    We evaluate our method on the physically-realistic cluttered bin-picking dataset SynPick and the YCB-Video dataset and demonstrate improved pose estimation accuracy as well as better object detection accuracy.
\end{abstract}


\section{Introduction}
Object detection is the task of localizing instances of object categories in images---typically by predicting bounding box parameters. 6D pose estimation aims at predicting the position and orientation of  objects in the sensor coordinate system.
Both tasks are essential for many autonomous robots and a prerequisite for object manipulation. 

 
Although single-view pose estimation models have made significant progress in recent years, they face difficulties in cluttered environments~\citep{hodavn2020bop} hampered by 
occlusions, reflective surfaces, transparency, and other challenges.
One way to address these challenges is to utilize a sequence of images of the scene instead of a single image.
In a video sequence, image features and object attributes evolve smoothly over time.
Models can benefit from imbuing image features and predictions from the previous frames
while processing the current frame.
Also, enforcing temporal consistency of the image features and pose predictions from consecutive frames can 
facilitate efficient learning and better accuracy.
Despite the apparent advantages of temporal processing, the popularity of single-view pose estimation methods can be attributed to the complexity, computation, and memory overhead of video pose estimation methods.
Furthermore, CNN-based models for video processing often utilize 3D convolutions, which need more parameters and are slow compared to their 2D counterparts.

\begin{figure}
    \centering
    \newlength{\imgre}
    \setlength{\imgre}{0.16\textwidth}
    \setlength{\tabcolsep}{0.01cm}
    \begin{tabular}{ccc}
     \includegraphics[width=\imgre]{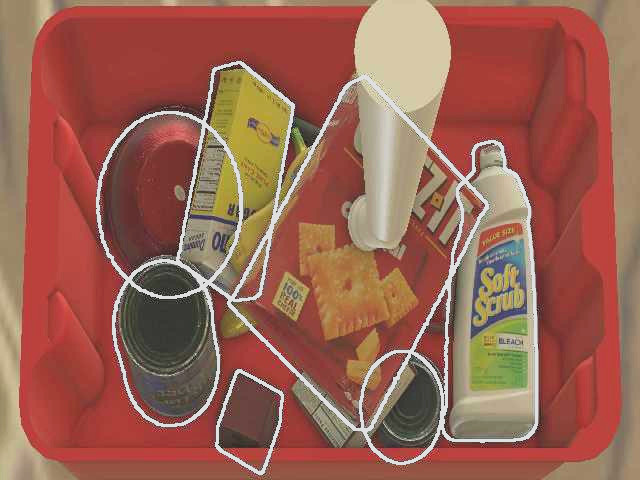} &
     \includegraphics[width=\imgre]{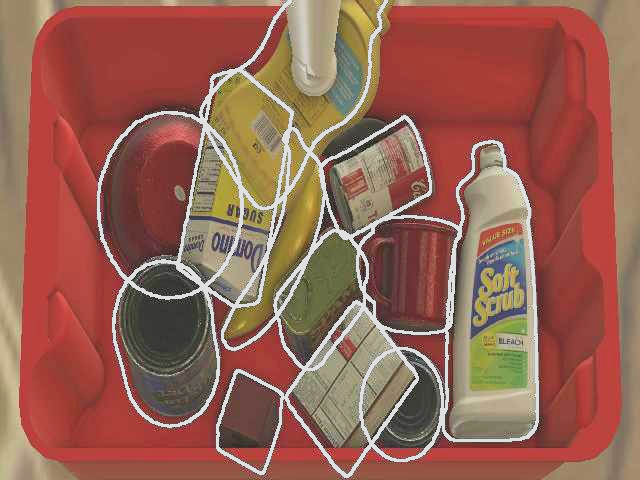} &
     \includegraphics[width=\imgre]{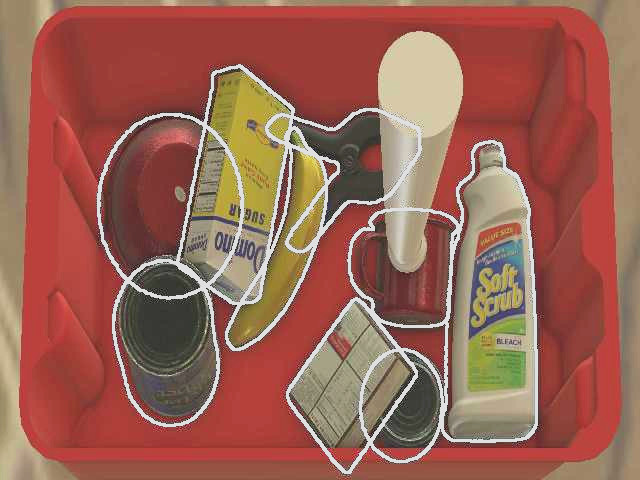} \\[-1.4ex]
      {\scriptsize (1)} & {\scriptsize (2)} & {\scriptsize (3)} \\
     \includegraphics[width=\imgre]{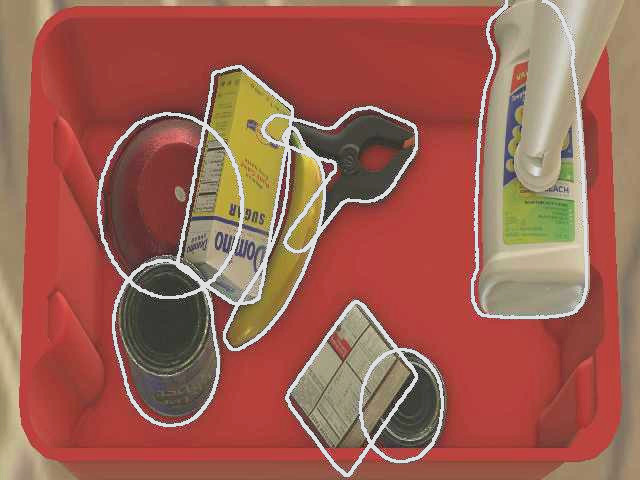} &
     \includegraphics[width=\imgre]{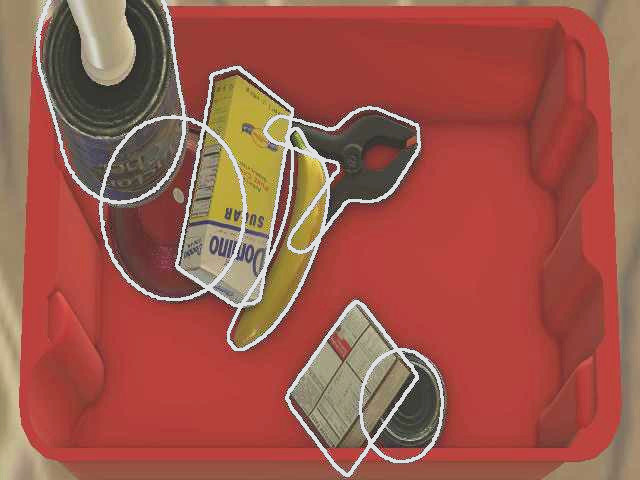} &
     \includegraphics[width=\imgre]{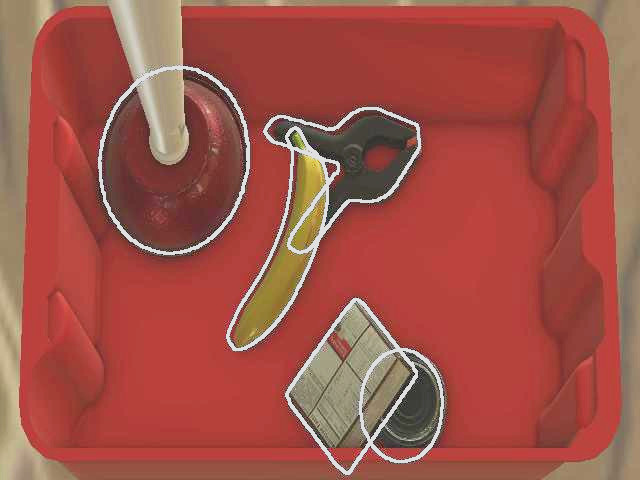} \\[-1.4ex]
     {\scriptsize (4)} & {\scriptsize (5)}& {\scriptsize (6)} \\
    \end{tabular}
    \caption{Multi-object pose predictions for a cluttered bin-picking scene from the SynPick dataset (Untargeted-pick, Sequence 38).
    MOTPose jointly detects and estimates 6D pose for all objects in the scene in a single step
    using a vision-transformer model by fusing temporal information across multiple frames.
    Predicted object poses are visualized by contours.}
    \label{fig:teaser}
    \vspace{-5mm}
\end{figure}

Lately, the multi-head attention-based transformer architecture, which was initially proposed for natural language processing tasks, has shown tremendous capabilities in modeling long-term dependencies in many domains like audio, image, video, etc.~\citep{han2022survey, wen2023transformers, khan2022survey, khan2022survey, Liu:EfficientViT:CVPR2023, Li:MaskDINO:CVPR2023}. 
Vision transformer architectures also enable single-stage models that jointly perform object detection and pose estimation for all objects in the scene in one forward pass~\citep{yolopose2022, arash2021gcpr}.
This ability is handy when dealing with highly cluttered bin-picking scenarios (see~\cref{fig:teaser}).
In this work, we propose a vision transformer model for multi-object 6D pose estimation from monocular video sequences.
The core component of the proposed MOTPose method is a cross-attention-based temporal fusion mechanism that fuses features from multiple past frames while processing the current frame. 
We use the stacked object embeddings from the past time steps as key and value in the cross-attention computation while the object embeddings from the current time step serve as query.
To counter the permutation-invariant nature of the attention mechanism in the temporal fusion modules, we utilize relative frame encoding (RFE).

\noindent Our contributions include:
\begin{itemize}
    \item a multi-object pose estimation model for dynamic video sequences,
    \item a method for cross-attention-based temporal fusion of object embeddings and object-specific outputs over multiple frames,
    \item SynPick-Ext, an extended version of the physically-realistic dataset SynPick consisting of 300 additional video sequences for each action split, and 
    \item quantitative evaluation of the joint object detection and pose estimation task on SynPick, and competitive results on YCB-Video while being lighter and faster than other methods.
\end{itemize}

\section{Related Work}

\subsubsection{Monocular Pose Estimation}
Object pose estimation from RGB images has been a long-standing problem in computer vision.
The traditional methods before the advent of deep learning include template-based methods~\citep{Hinterstoier2012GradientRM, Hinterstoier2012ModelBT} and 
feature-based methods~\citep{Rothganger20063DOM,Pavlakos20176DoFOP,Tulsiani2014ViewpointsAK}.
Modern deep-learning-based approaches include direct methods that regress the 6D pose parameters given the input RGB image~\citep{xiang2017posecnn,periyasamy2018pose,Wang_2021_GDRN,di2021so},
keypoint-based methods that predict the pixel coordinates of 3D keypoints first and then use the \mbox{\textit{perspective-n-points}} (P\textit{n}P) algorithm to recover 6D pose~\citep{rad2017bb8,tekin2018real,hu2019segmentation,peng2019pvnet,hu2020single}, and refinement-based methods.
The latter iteratively refine an initial pose estimate using either the \textit{render-and-compare} framework~\citep{li2018deepim,manhardt2018deep,labbe2020,periyasamy2019refining, castro2023crt}
or optical flow~\citep{hai2023shape, hu2022perspective}.
Most monocular pose estimation methods are multi-staged. The standard pipeline involves object detection and/or semantic segmentation, target object crop extraction, and pose estimation from the extracted crop. To enable end-to-end trainable multi-staged models, specialized operations 
like~\textit{non-maximum suppression} (NMS),~\textit{region-of-interest pooling} (ROI), or~\textit{anchor boxes} are employed.
Notable single-stage methods include~\citep{Capellen2020, hu2019segmentation, pyrapose2021}.
Our proposed MOTPose method also incorporates single-stage design elements in its architecture. 

\subsubsection{Pose Estimation as Set Prediction}
In recent years, vision transformer architectures, that formulate computer vision tasks like object detection, instance segmentation, and pose estimation as a set prediction problem, are achieving impressive results.
\citet{carion2020end} introduced DETR, the pioneering work in this new class of methods.
Several methods extended DETR for multi-object pose estimation~\cite{yolopose2022,arash2021gcpr, jantos2023poet}.
Following these methods, the proposed MOTPose model formulates multi-object pose estimation from video sequences as a set prediction problem.

\subsubsection{6D Pose Tracking}
Many of the early works for 6D pose tracking were based on particle filtering~\citep{Azad20116DoFMT,Pauwels2013RealTimeMR,yu2014tracking}, but
the performance of particle filters heavily depends on the accuracy of the observation model.
\citet{poserbpf2019} introduced PoseRBPF utilizing a CNN-based observation model in the particle filtering framework.
\citet{wen2020se} introduced \textit{se}(3)-TrackNet, which achieved state-of-the-art-results in object pose tracking from RGB-D images.
In contrast to \textit{se}(3)-TrackNet, our MOTPose method only needs RGB input and can estimate 6D pose for all objects in the input images in one stage.

\subsubsection{Multi-Object Tracking}
Multi-object tracking aims at tracking 2D bounding boxes of the target instances in a given video sequence. The task is often challenging, due to the presence of multiple instances of the same category.
To address the problem of matching detections and tracked objects, sophisticated matching strategies were proposed~\citep{tracktor_2019_ICCV, zhou2020tracking, zhou2020tracking, Xu2021TransCenterTW}.
In this work, we focus mainly on improving pose estimation accuracy by fusing information over multiple time steps. 
Thus, instead of focusing on the tracking metrics, we emphasize the standard pose estimation metrics---ADD-S and \mbox{ADD(-S)}---discussed in \cref{sec:metric}.

\subsubsection{Tracking-by-Attention in DETR-Like Models}
Recently, \citet{trackformer2022} proposed TrackFormer, by introducing the \textit{tracking-by-attention} framework in a DETR-like architecture.
Their key idea is to use object embeddings from time step $t$ as object queries in time step $t$+1.
Propagating object embeddings over multiple time steps enables tracking the object over a long video sequence.
State-of-the-art methods for multi-object tracking utilizing the tracking-by-attention framework include MOTR~\citep{motr2022} and TransTrack~\citep{sun2020transtrack}.
The main downside of such methods is that the number of object queries in a time step is dynamic, which makes efficient vectorized implementation harder and results in a slow training process.
In contrast to the \textit{tracking-by-attention} framework, in our model, we fuse a fixed set of object embeddings and object-specific outputs from multiple time steps using cross-attention-based modules.

\begin{figure*}
    \centering
    \includegraphics[width=\linewidth]{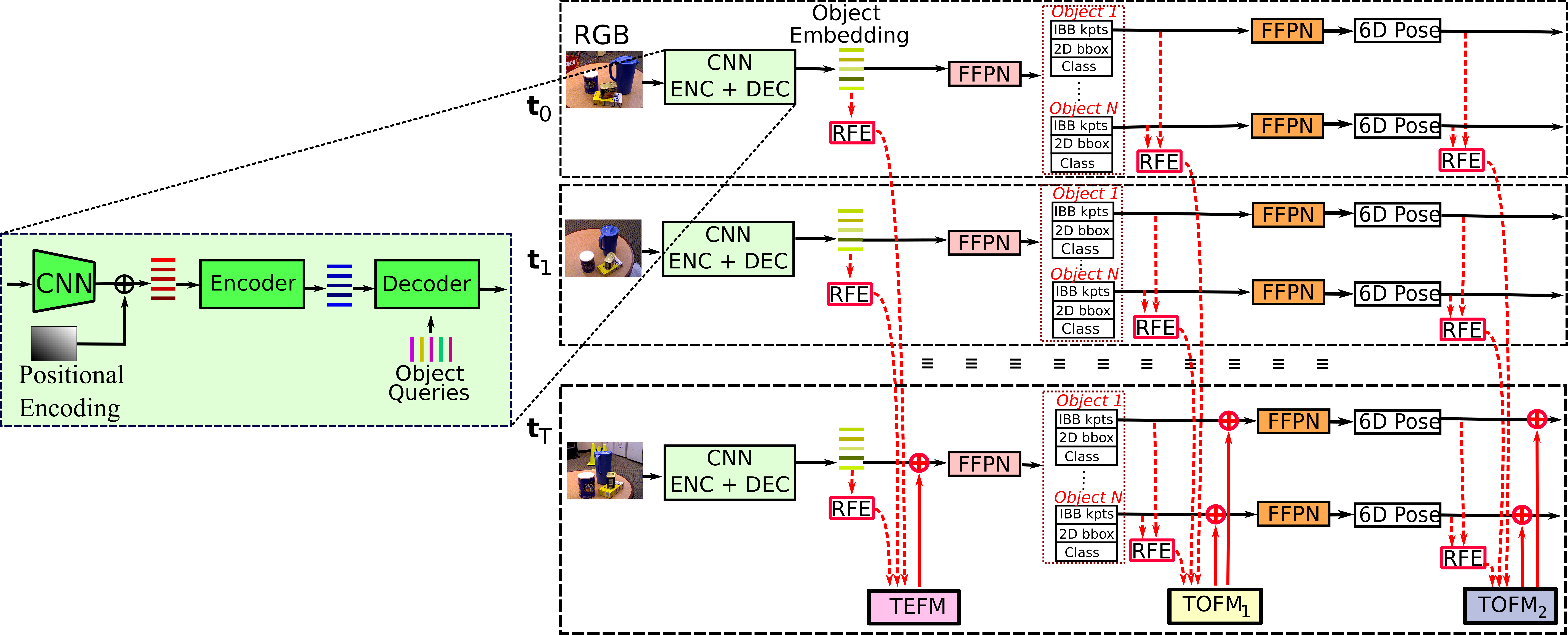}
    \caption{MOTPose architecture. 
    Positional Encoding: pixel coordinates represented using sine and cosine functions of different frequencies.
    Object Queries: learned embeddings that are trained jointly with the model and remain fixed during inference (Sec.~\ref{sec:yolopose}).
    FFPN: Feed Forward Prediction Network.
    TEFM: Temporal Embedding Fusion Module (Sec.~\ref{sec:TEFM}, Fig~\ref{fig:tffm}).
    TOFM: Temporal Object Fusion Module (Sec.~\ref{sec:TOFM}).
    $\oplus$: Element-wise addition.
    $ \color{red} \oplus $: Residual connection.
    The dashed red lines represent temporal connections. 
    All modules that share a color also share weights.
    At each time step, object embeddings are generated using a CNN backbone and transformer-based encoder-decoder modules.
    The image features from the backbone are augmented with positional encoding.
    The object embeddings are processed in parallel using FFPNs to generate class probability, bounding box, and 6D pose parameters. At time step $t_T$, the object embeddings of different time steps are fused using TEFM.
    Similarly, object-specific predictions like the keypoints and the 6D pose parameters of different time steps are fused using TOFM. 
    While fusing object embeddings and object-specific outputs from different time steps, Relative Frame Encoding (RFE) is added element-wise to uniquely identify the respective time step.}
    \label{fig:MOTPose}
\end{figure*}


\section{Method}


\subsection{Multi-Object Pose Estimation as Set Prediction}
\label{sec:yolopose}

Following YOLOPose~\citep{yolopose2022}, we formulate multi-object pose estimation as a set prediction problem.
YOLOPose exploits the permutation-invariant nature of the attention mechanism to generate a set of tuples---each consisting of class probabilities,
2D bounding box, 3D bounding box, position and orientation parameters.
The 3D bounding box parameters are represented using the interpolated bounding box (IBB) keypoints~\citep{Li_2021_CVPR}.
YOLOPose employs a ResNet backbone for feature extraction (CNN). Positional encoding compensates for the loss of spatial information in the permutation-invariant attention computation. Combined image features and positional encodings are provided to the encoder module, which uses the multi-head self-attention mechanism to generate encoder feature embeddings. In the decoder, the cross-attention mechanism is employed between the encoder feature embeddings and a set of $N$ learned embeddings called object queries to generate $N$ object embeddings, which are then processed by feed-forward prediction networks (FFPNs) to generate class probabilities, 2D bounding box, and IBB keypoints in parallel. 
The IBB keypoints are then processed by a subsequent FFPN to estimate the translation and rotation parameters.
Since the cardinality of the predicted set is fixed, the model is trained to predict \text{\O} classes after detecting all the target objects present in the image.
By associating predictions and ground truth objects using a bipartite matching algorithm~\cite{kuhn1955hungarian}, YOLOPose is trained end-to-end.


\subsection{MOTPose Architecture}
The architecture of the proposed MOTPose model is shown in~\cref{fig:MOTPose}.
We base the single-frame processing of MOTPose on the YOLOPose model.
The transformer-based encoder-decoder modules generate object embeddings of cardinality $N$ from CNN-computed
image features that are augmented with positional encoding.
FFPNs process the object embeddings to generate object-specific outputs.
The object embeddings and the object-specific outputs from the past time steps provide rich temporal information that
can be leveraged while processing the current frame.
To this end, we fuse object embeddings and object-specific outputs from multiple past time steps using the Temporal Embedding Fusion Module (TEFM, Sec.~\ref{sec:TEFM}) and the Temporal Object Fusion Module (TOFM, Sec.~\ref{sec:TOFM}), respectively, before generating outputs for the current time step. 
To enable the fusion of embeddings and object parameters over multiple time steps using the permutation-invariant attention mechanism, we utilize \emph{relative frame encoding} (RFE), which encodes the number of time steps relative to the current frame using 1D sinusoidal functions.


\subsubsection{Temporal Embedding Fusion Module (TEFM)}
\label{sec:TEFM}

\begin{figure}
    \centering
    \includegraphics[width=\linewidth, height=3cm]{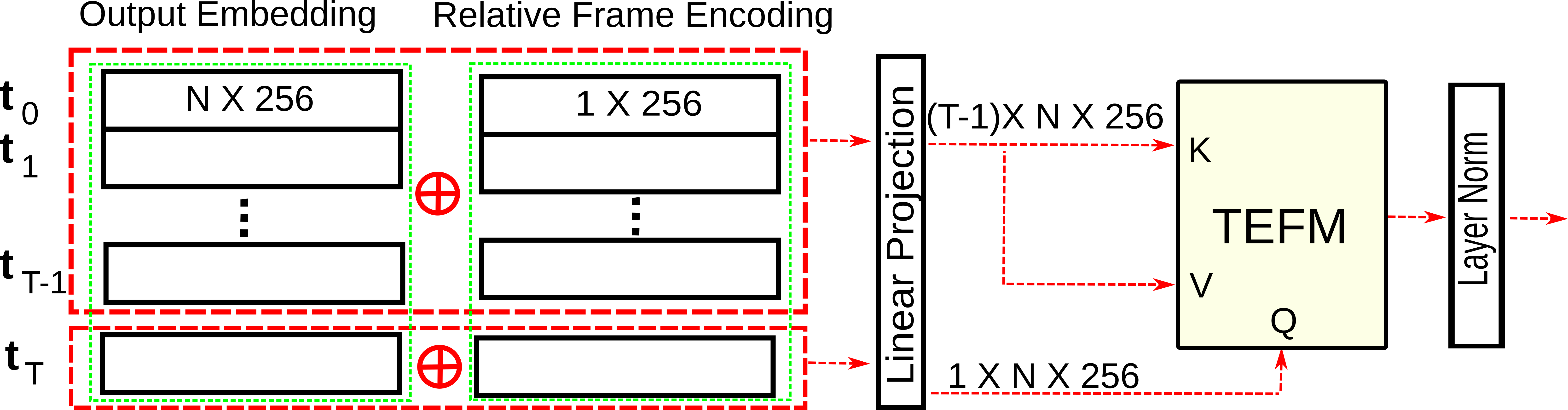}
    \caption{Temporal Embedding Fusion Module (TEFM). 
    $ \color{red} \oplus $: Concatenation operation.
    The object embeddings at each time step of shape N${\times}$256 are added element-wise with relative frame encoding (RFE).
    The resulting vectors for time steps $t_0-t_{T{-}1}$ are stacked to form \textit{key} as well as \textit{value} for the cross-attention operation in TEFM,
    whereas the embedding at time step $T$ acts as \textit{query}.
    \label{fig:tffm}
    }
    \vspace{-5mm}
\end{figure}

At each time step, the decoder generates object embeddings of shape $N\!\!\times\!256$, where $N$ is the cardinality of the object set to be predicted.
TEFM, shown in~\cref{fig:tffm}, fuses object embeddings from multiple time steps to extract valuable temporal information. 
First, relative frame encoding is concatenated with the object embeddings, and then the resulting embeddings are projected back to 256 dimensions using linear layers.
The stacked embeddings until $T{-}1$ time steps form \textit{key} and \textit{value} for the cross-attention operation in TEFM, whereas the embedding from the time step $T$
is used as \textit{query}. 
This allows the object embeddings from time step $T$ to interact with object embeddings from all previous time steps.
The key-query similarity is reflected in the resulting attention weights.
These attention weights are used to weigh the \textit{value} vectors, which in our case are the object embeddings from all previous time steps.
After applying layer normalization, the output of TEFM is added element-wise to the object embeddings of time step $T$, representing a residual connection.

\subsubsection{Temporal Object Fusion Module (TOFM)}
\label{sec:TOFM}
In addition to fusing embeddings using TEFM, we employ two TOFM modules to fuse object-specific outputs. 
The design of TOFM is similar to that of TEFM, except for the usage of additional linear projection layers at the beginning and the end. The object embeddings are of shape 
$N\!\!\times\!256$, whereas the shape of the predictions is $N{\times}P$, which depends on the prediction generated;
three in the case of translation prediction, six in the case of rotation prediction, and 32 in the case of keypoints. 
We use a linear layer to project the predictions to a 256-dimensional vector and supplement them with RFEs.
After computing cross-attention, we project the resulting embeddings back to $N{\times}P$.
TOFM$_1$ is used for fusing keypoints and TOFM$_2$ is used for fusing pose parameters.




\subsection{Matching}
\label{sec:match}
We use the bipartite matching algorithm~\cite{carion2020end,yolopose2022,kuhn1955hungarian} to associate predicted and ground-truth objects.
 Despite jointly estimating 2D bounding box, class probabilities, key points, and pose parameters, similar to~\citep{yolopose2022, periyasamy2023yolopose}, 
we use only the bounding box and the class probability components in the matching cost function. This is based on the empirical observation that a combination of the bounding box and the class probability components alone is enough to ensure an optimal match between the ground-truth and the predicted sets.

\subsection{Loss Function}
\label{sec:loss_fns}
The \textit{Hungarian loss} used to train MOTPose is a weighted combination of five components:

\subsubsection{Class Probability Loss}
We use the standard negative log-likelihood (NLL) loss to train the classification branch of the model.
To deal with the class imbalance due to the \text{\O} class appearing disproportionately often, we weigh it down by a factor of 0.1.

\subsubsection{Bounding Box Loss}
To train the bound box prediction branch of our model, we employ a linear combination of the generalized \mbox{IOU~\citep{rezatofighi2019generalized}} and the $\ell_1$-loss.

\subsubsection{Keypoint Loss}
\label{sec:kploss}
We use a weighted combination of the $\ell_1$-loss and the cross-ratio consistency loss~\citep{Li_2021_CVPR,yolopose2022} to train the keypoint estimation branch.



\subsubsection{Pose Loss}
We decouple the pose loss into a translation and a rotation component. For translation, we employ the $\ell_2$-loss.
For rotation, we use the symmetry-aware ShapeMatch-loss proposed by~\citet{xiang2017posecnn}.


\subsubsection{Temporal Consistency Loss}
 We enforce temporal consistency using the $\ell_2$-loss between the object embeddings of consecutive time steps.
Embeddings evolve smoothly over frames and any big changes are undesirable.
Thus, the $\ell_2$-loss, which penalizes bigger differences significantly more than smaller differences, is a natural choice.

\section{Evaluation}

\subsection{Datasets}

\subsubsection{YCB-Video}
We use the challenging YCB-Video dataset~\citep{xiang2017posecnn} to benchmark the performance of our model against other state-of-the-art methods.
The dataset consists of 92 (80 training and 12 testing) moving-camera video sequences of static scenes with multiple objects.
High-resolution 3D models of all 21 objects are provided with the dataset.
Following~\citet{poserbpf2019, li2018deepim}, we use all the frames in the test split for evaluation.
Additionally, we utilize the synthetic dataset provided by Xiang et al.~\citep{xiang2017posecnn} to train our model.

\subsubsection{SynPick}
SynPick~\citep{periyasamy2021synpick} is a physically-realistic synthetic dataset of dynamic bin-picking scenes that contain a chaotic pile of the same 21 YCB-Video objects in a tote.
It consists of simulations of three different bin-picking actions: move, targeted pick, and untargeted pick.
For each action, SynPick provides 300 video sequences: 240 for training and 60 for testing.
Moreover, the dataset generator is publicly available\footnote{\url{https://github.com/AIS-Bonn/synpick}} making it easy to generate additional data, if needed.
In contrast to the commonly used object pose estimation datasets~\citep{Linemodoccluded, Hinterstoier2012ModelBT, hodavn2020bop}, which consist of static tabletop scenes with a relatively low degree of occlusion, SynPick is highly cluttered and the gripper movements generate complex object interactions.
Moreover, the objects in the SynPick dataset appear in a wide range of pose configurations and multiple instances of the same object are present in the scenes.
Thus, SynPick is an ideal dataset for evaluating the proposed MOTPose model.

\subsection{Metrics}
\label{sec:metric}

We report the area under the curve (AUC) of the ADD and ADD-S metrics at an accuracy threshold of 0.1m for non-symmetric and symmetric objects, respectively~\citep{xiang2017posecnn}.
The ADD metric is the average $\ell_2$ distance between the subsampled mesh points in the ground truth and the predicted pose, 
whereas the symmetry-aware ADD-S metric is the average distance between the closest subsampled mesh points in the ground truth and the predicted pose.
The \mbox{ADD(-S)} metric combines both ADD and ADD-S into one metric by utilizing ADD for objects without symmetry and ADD-S for objects exhibiting symmetry.



\begin{figure*}[t]
    \centering \footnotesize
    \newlength{\imgres}
    \setlength{\imgres}{0.2\textwidth}
    \setlength{\tabcolsep}{0.01cm}
    \begin{tabular}{ccccc}
     \includegraphics[width=\imgres]{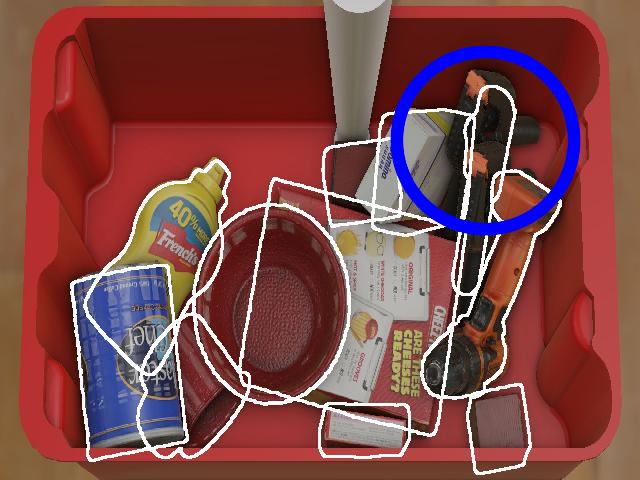} &
     \includegraphics[width=\imgres]{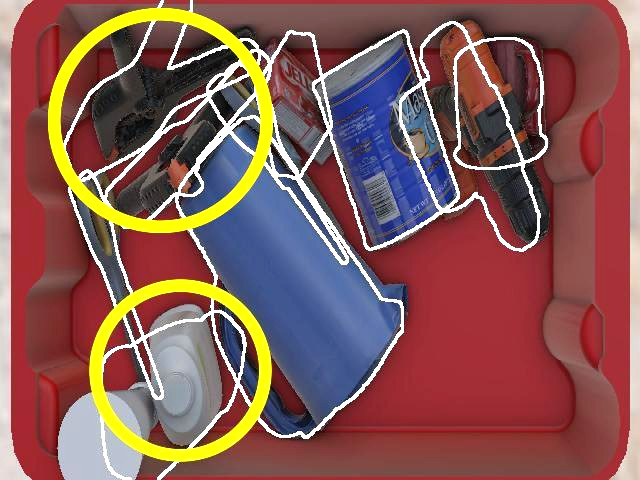} &
     \includegraphics[width=\imgres]{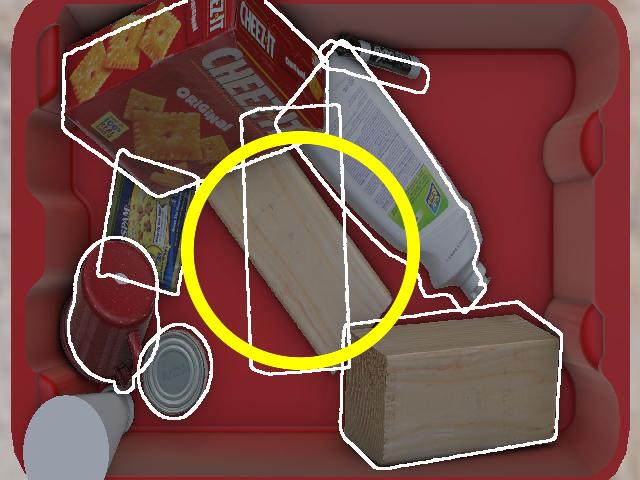} &
     \includegraphics[width=\imgres]{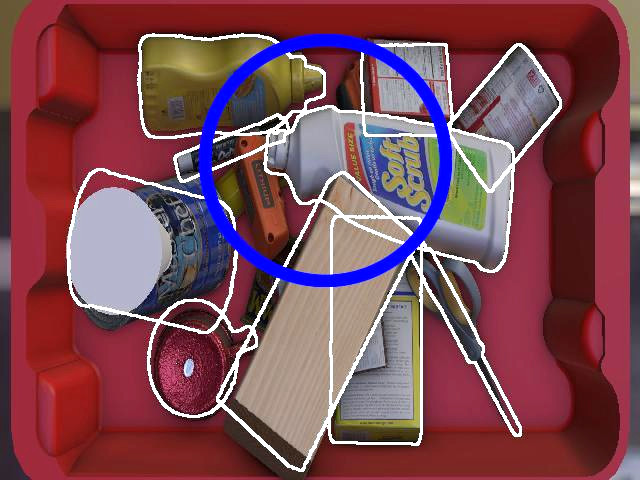} &
     \includegraphics[width=\imgres]{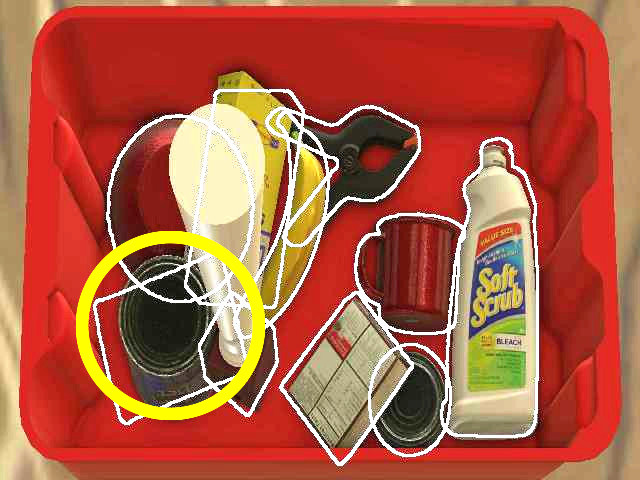} \\
        
     \includegraphics[width=\imgres]{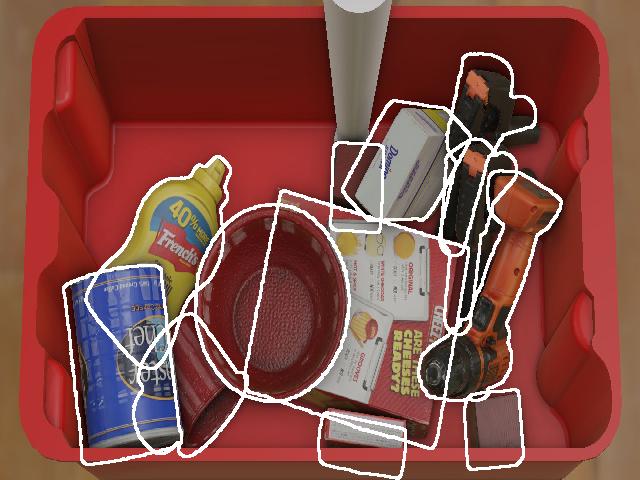} &
     \includegraphics[width=\imgres]{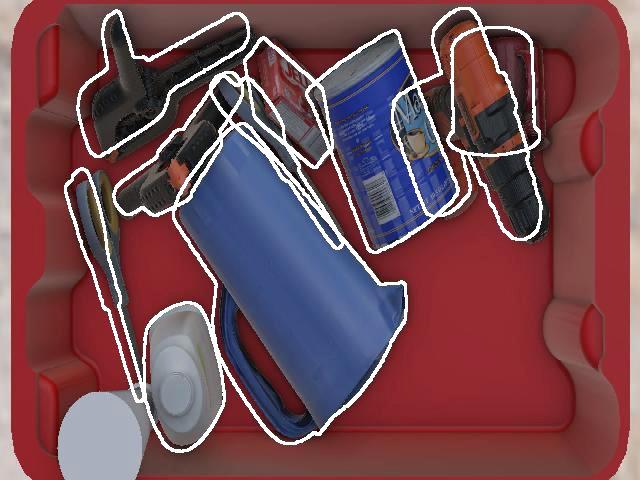} &
     \includegraphics[width=\imgres]{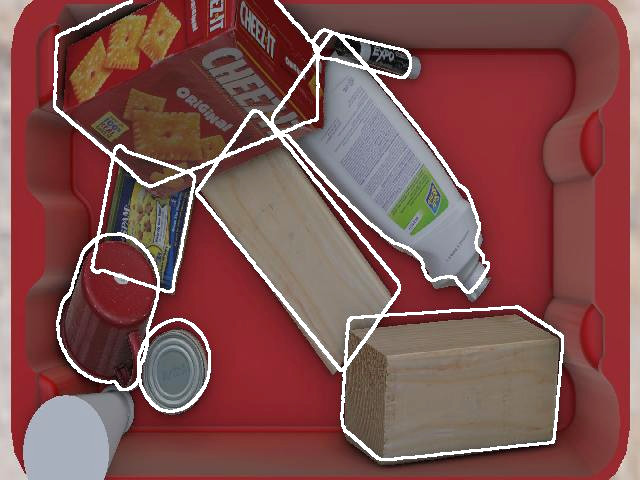} &
     \includegraphics[width=\imgres]{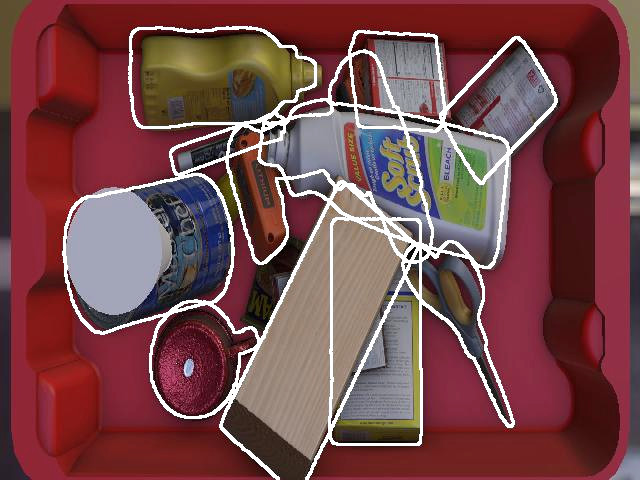} &
     \includegraphics[width=\imgres]{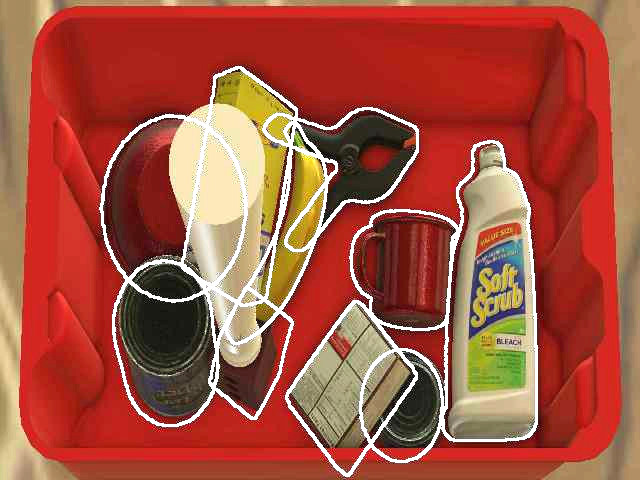} \\
     (a) & (b) & (c) & (d) & (e) \vspace*{-1mm}
    \end{tabular} 
    \caption{Qualitative results on SynPick. 
    6D pose predictions are visualized by object contours.
    Top: Predictions from the model without temporal fusion.
    Bottom: Predictions from the model with temporal fusion.
    Temporal fusion facilitates better pose prediction as well as object detection accuracies.
    The blue circles highlight failed object detections and the yellow circles highlight erroneous pose predictions.}
    \label{fig:result}
\end{figure*}

\begin{table}
\centering
\begin{threeparttable}
\scriptsize
\caption{Quantitative results on the SynPick dataset.}
\label{tab:synpick-details}
\setlength{\aboverulesep}{0pt}
\setlength{\belowrulesep}{0pt}
\setlength\tabcolsep{1pt}
{
\begin{tabular}{c|cc|cc||cc|cc}
\toprule
 & \multicolumn{4}{c||}{\thead{MOTPose without\\ Temporal Fusion}}  & \multicolumn{4}{c}{\thead{ MOTPose with\\ Temporal Fusion}}  \\
\midrule
\thead{Obj.~~~\\ID$^{\dagger}$}& 
\thead{\scriptsize AUC\,of \\ \scriptsize ADD-S} & \thead{ \scriptsize AUC\,of \\ \scriptsize ADD(-S)} &
\thead{\scriptsize AUC\,of \\ \scriptsize ADD-S \\ \scriptsize{@}0.1d} & \thead{\scriptsize AUC\,of \\  \scriptsize ADD(-S) \\ \scriptsize{@}0.1d} & 
\thead{\scriptsize AUC\,of \\ \scriptsize ADD-S} & \thead{\scriptsize AUC\,of \\ \scriptsize ADD(-S)} & 
\thead{\scriptsize AUC\,of \\ \scriptsize ADD-S \\\scriptsize{@}0.1d} & \thead{\scriptsize AUC\,of \\ \scriptsize ADD(-S) \\\scriptsize{@}0.1d} \\
\midrule

1              & 88.8 & 72.2   & 86.1 &  53.4    &  88.5 & 79.1 & 86.8 & 61.2             \\
2              & 90.7 & 82.5   & 89.5 &  76.2    &  91.4 & 84.2 & 90.2 & 78.4             \\
3              & 80.8 & 74.4   & 79.1 &  63.9    &  81.6 & 76.2 & 80.2 & 69.5             \\
4              & 72.5 & 64.1   & 70.1 &  43.9    &  73.5 & 68.0 & 71.2 & 45.1             \\
5              & 80.3 & 72.2   & 78.8 &  62.3    &  80.2 & 74.8 & 78.9 & 67.9             \\
6              & 81.1 & 64.1   & 68.1 &  19.1    &  81.8 & 75.1 & 72.2 & 25.6             \\
7              & 69.9 & 63.4   & 66.3 &  48.2    &  70.9 & 65.7 & 68.4 & 48.3             \\
8              & 65.8 & 60.3   & 60.6 &  40.9    &  67.4 & 62.1 & 63.3 & 32.0             \\
9              & 84.3 & 76.1   & 80.6 &  56.4    &  85.1 & 78.9 & 82.5 & 56.5             \\
10             & 78.0 & 70.5   & 73.9 &  56.9    &  80.3 & 73.9 & 77.9 & 64.9             \\
11             & 92.8 & 84.7   & 92.2 &  79.4    &  93.1 & 85.8 & 92.4 & 81.8             \\
12             & 85.7 & 76.9   & 85.2 &  71.1    &  87.0 & 80.7 & 86.4 & 76.9             \\
13$^*$         & 89.0 & 89.0   & 83.2 &  83.2    &  89.5 & 89.5 & 85.9 & 85.9             \\
14             & 84.9 & 74.8   & 80.8 &  49.0    &  85.9 & 78.5 & 82.5 & 45.6             \\
15             & 90.5 & 83.7   & 89.9 &  75.2    &  92.9 & 87.2 & 92.3 & 83.0             \\
16$^*$         & 90.0 & 90.0   & 88.8 &  88.8    &  90.0 & 90.0 & 88.9 & 88.9             \\
17             & 72.0 & 65.0   & 65.1 &  49.7    &  75.9 & 69.5 & 71.1 & 55.2             \\
18             & 68.1 & 62.4   & 61.7 &  36.6    &  66.9 & 61.9 & 60.4 & 36.2             \\
19$^*$         & 76.0 & 76.0   & 73.6 &  73.6    &  79.0 & 79.0 & 77.5 & 77.5             \\
20$^*$         & 80.5 & 80.5   & 75.7 &  75.7    &  83.6 & 83.6 & 81.8 & 81.8             \\
21$^*$         & 75.9 & 75.9   & 72.2 &  72.2    &  76.3 & 76.3 & 69.7 & 69.7             \\  
\midrule                               
\footnotesize \rule{0mm}{2.6mm} Mean~          & \footnotesize 80.8 & \footnotesize 74.2   & \footnotesize 77.2 &  \footnotesize 60.8    &  \footnotesize \textbf{82.0} & \footnotesize \textbf{77.1} & \footnotesize \textbf{79.1} & \footnotesize \textbf{63.4}             \\
\bottomrule
\end{tabular}
}
\begin{tablenotes}\scriptsize
     \item $^*$ Symmetric objects.
     \item $\dagger$ Object ID in the standard order of YCB-Video.
\end{tablenotes}
\end{threeparttable}

\end{table}

\begin{table}
    \centering
    \caption{Cardinality Error on SynPick Splits [${\times} 10 ^{-2}$].}
  \label{tab:cardinality}
    \footnotesize
    \setlength{\aboverulesep}{1pt}
    \setlength{\belowrulesep}{1pt}    
\vspace{-3mm}
\begin{tabular}{l|c|c|c|c}
  \toprule
  Method& \thead{Move} & \thead{Targeted\\pick} & \thead{Untargeted \\ pick} & All \\
  \bottomrule
  W\slash o temporal fusion & 3.26 & 1.64 & 0.48 & 2.06  \\
  With temporal fusion & \textbf{0.62} & \textbf{0.52} & \textbf{0.44} & \textbf{0.53} \\
  \bottomrule
\end{tabular}
\end{table}

\begin{table}
    \centering
    \caption{False negative detections on SynPick Splits [${\times} 10 ^{-2}$].}
  \label{tab:fn_error}
    \footnotesize
    \setlength{\aboverulesep}{1pt}
    \setlength{\belowrulesep}{1pt}
\vspace{-3mm}
\begin{tabular}{l|c|c|c|c}
  \toprule
  Method& \thead{Move} & \thead{Targeted\\pick} & \thead{Untargeted \\ pick} & All \\
  \bottomrule
  W\slash o temporal fusion & 2.79 & 1.39 & 1.36 & 1.79  \\
  With temporal fusion & \textbf{0.57} & \textbf{0.44} & \textbf{0.44} & \textbf{0.48} \\
  \bottomrule
\end{tabular}
\end{table}

\subsection{Implementation Details}
\label{sec:imp_details}
Following~\citep{carion2020end, periyasamy2023yolopose}, we choose the cardinality of the predicted set $N$ proportional to
the maximum number of objects in an image in the respective datasets: 30 for SynPick and 20 for YCB-Video.
In ~\cref{sec:loss_fns}, the bounding box components are weighted using factors 2 and 5, and the keypoint components are weighted with factors 10 and 1.
The pose component and the temporal consistency component are weighted down using factors 0.05 and 0.1, respectively.
The encoder and decoder modules consist of six layers each.
All the embeddings used in our model are of dimension 256.
We train our model for 150 epochs using the AdamW optimizer with a learning rate of $1{\times}10^{-4}$ and early stopping.
We set the number of time steps $T$ to eight in the temporal fusion modules and use a batch size of 32 (four groups of eight consecutive images).
\subsection{Results on SynPick}

Formulating multi-object pose estimation as a set prediction problem enables joint object detection and pose estimation of all objects in the scene.
However, it compounds the size of the dataset required to train transformer models. Thus, to complement the existing 240 videos for training, we generate additional 300 video sequences for each action split.
We call this extended version \mbox{SynPick-Ext} and make it publicly available\footnote{\url{https://www.ais.uni-bonn.de/videos/tempose}}. 
We downsample the image resolution to 640${\times}$480.
SynPick consists of objects piled up in a tote and in many cases, objects are completely occluded.
To exclude heavily occluded objects, we use a minimum visibility threshold of 30\% in our evaluation.
In~\cref{fig:result}, we present pose estimates generated by our model with and without temporal fusion.
Both models generate predictions of admissible quality. However, the model without temporal fusion suffers from failed object detections (\cref{fig:result}(a),\,(d)), and
isolated highly erroneous pose predictions (\cref{fig:result}(b),\,(c),\,(e)). Temporal fusion helps in alleviating these shortcomings. 

In~\cref{tab:synpick-details}, we report quantitative results of our model. MOTPose achieves impressive AUC\,of\,ADD-S and AUC\,of\,ADD(-S) scores of 82.0 and 77.1, respectively, which is 
an improvement of 1.2 and 2.9 compared to the model without temporal fusion. Additionally, we also report the AUC metrics with a threshold of 10\% of the object diameter (AUC{@}0.1d).
This metric takes the object size into account better. In terms of  AUC\,of\,ADD-S and \mbox{ADD(-S){@}0.1d}, temporal fusion boosts the accuracy by 1.9 and 2.6 points, respectively.

Furthermore, to understand the impact of temporal fusion on object detection, we analyze the cardinality error and the bounding box accuracy metrics. The cardinality error is the difference between elements in the ground-truth and predicted sets.
Formally, given the ground-truth set $\mathcal{Y}$ and the predicted set $\hat{\mathcal{Y}}$, the cardinality error ($\text{CE}$) is defined as:
\begin{equation}\label{eqn:CE}
    \text{CE} = \frac{|(\mathcal{Y} - \hat{\mathcal{Y}}) \cup (\hat{\mathcal{Y}} - \mathcal{Y})|}{|\mathcal{Y}|}.
\end{equation}

In~\cref{tab:cardinality}, we report the cardinality error of our model on different splits of the SynPick dataset.
Over the complete test set, the cardinality error of the model without temporal fusion is 0.021, whereas it is only 0.005 for the model with temporal fusion.
The difference is more evident in the {\em Move} split, which is more challenging than the other two splits.

Although $\text{CE}$ reflects the set prediction ability of a model, in real-world bin-picking systems, the identity of the objects present in the bin might be known a priori~\citep{schwarz2018fast, schwarz2017nimbro}.
Thus, in this \emph{informed detection} scenario, false positives can be easily mitigated, whereas false negatives (FN), i.e., $|(\mathcal{Y} - \hat{\mathcal{Y}})| \slash |\mathcal{Y}|$ are detrimental.
In~\cref{tab:fn_error}, we report the false negatives of object detection. Over the entire test set, the model without temporal fusion has a FN rate of 0.018; with temporal fusion, the FN rate drops to 0.005.

To compare the bounding box detection accuracy, we analyze the average precision and recall metrics defined by COCO evaluation protocol\footnote{\url{https://cocodataset.org/\#detection-eval}}.
In~\cref{tab:bbox}, we report the AP@[IoU=0.50:0.95], AP@[IoU=0.50], AP@[IoU=0.75], and AR@[IoU=0.50:0.95] metrics of the models with and without temporal fusion.
Across all the reported metrics, temporal fusion yields consistent improvements.

\begin{table}
\begin{threeparttable}
    \centering
    
    \caption{Bounding Box Prediction Accuracy.}
  \label{tab:bbox}
    \footnotesize
    \setlength{\aboverulesep}{0pt}
    \setlength{\belowrulesep}{0pt}
      
\begin{tabular}{l|c|c|c|c}
  \toprule
  Method&  \thead{AP$^{\dagger}$}  & \thead{AP@\\ {[IoU=0.50]} } & \thead{AP@\\ {[IoU=0.75]} } & \thead{AR$^{\dagger}$} \\
  \midrule
  W\slash o temporal fusion & 0.756 & 0.872 & 0.853 & 0.789  \\
  With temporal fusion & \textbf{0.779} & \textbf{0.876} & \textbf{0.858} & \textbf{0.811} \\
  \bottomrule
\end{tabular}
\begin{tablenotes}\footnotesize
    \item$^{\dagger}$ @[IoU=0.50:0.95]
\end{tablenotes}
\end{threeparttable}
\vspace{-3mm}
\end{table}


\subsection{Results on YCB-Video}
\label{sec:results_ycbv}

In~\cref{tab:ycbv-details}, we report the quantitative comparison of our MOTPose model against state-of-the-art methods on the YCB-Video dataset.
In our experiments, we fuse seven previous frames ($T{=}8$) in MOTPose. Since our model does not produce outputs for the initial $T{-}1$ frames in a video sequence, we report the accuracy scores excluding the initial frames.
Temporal fusion enables considerable improvement in the MOTPose model: 0.9 and 1.3 accuracy points in terms of the AUC\,of\,ADD-S and AUC\,of\,ADD-(S) metric, respectively.
Compared to DeepIM-Tracking~\citep{li2018deepim}, our method achieves a comparable AUC\,of\,ADD-S score and a slightly worse 
AUC\,of\,ADD-(S) score.
DeepIM-Tracking formulates 6D pose tracking as pose refinement, i.e., pose prediction from the previous frame is used to initialize the \text{render-and-compare} pose refinement for the current step. To initialize the first frame, the authors used the ground-truth pose.
While~\citet{castro2023crt} achieve a significantly better accuracy than MOTPose, they perform only pose refinement.
In contrast, our method performs multi-object detection and pose estimation jointly.
Moreover, MOTPose accuracy is comparable to the state-of-the-art multi-object pose estimation method of \citet{periyasamy2023efficient}
in terms of the AUC\,of\,ADD-(S) metric and only slightly worse in terms of the AUC\,of\,ADD-S metric.
Note that the frame rates reported in~\cref{tab:ycbv-details} are observed on GPUs of different generations and
the values are provided only for a relative comparison.


%
%

\subsection{Ablation Study}
\label{sec:ablation}
To understand the contribution of the individual components to the overall performance of MOTPose, we investigated removing different components of the model and varying the number of time steps used in the fusion modules.
In~\cref{tab:ablation}, we report the results of the ablation experiment on SynPick. Removing the TEFM module resulted in a big drop in the overall accuracy of the MOTPose model. 
In terms of the AUC\,of\,\mbox{ADD(-S)} metric, the MOTPose model without the TEFM module achieves a score of 74.9, compared to 77.1, while the AUC\,of\,ADD-S metric score drops by 0.6.
Similarly, removing the TOFM module results in a drop of 0.9 AUC\,of\,\mbox{ADD(-S)} and 1.8 AUC\,of\,ADD-S accuracy scores.
Moreover, in terms of the number of time steps used in the fusion modules, eight time steps resulted in the best performance overall.

\begin{table}
  \centering
  \caption{Results on the YCB-Video dataset.}
\label{tab:ycbv-details}
  \footnotesize
  \setlength{\aboverulesep}{0pt}
  \setlength{\belowrulesep}{0pt}
\vspace{-3mm}  
\begin{tabular}{l|c|c|c}
\toprule
Method  & \thead{AUC of \\ ADD-S} & \thead{AUC of \\ ADD(-S)} & \textit{fps}\\
\midrule

CRT-6D~\citep{castro2023crt} & - & \textbf{87.5} & 30 \\
\citet{periyasamy2023efficient} & \textbf{92.0} & 84.7 & 26 \\
DeepIM-Tracking~\citep{li2018deepim} & 91.0 & 85.9 & 13 \\
MOTPose w\slash o temporal fusion & 90.3 & 83.2 & 59 \\
MOTPose with temporal fusion & 91.2 & 84.5 & 30 \\
\bottomrule
\end{tabular}
\end{table}

\begin{table}
      \centering
      \caption{Ablation study results on the SynPick dataset.}
    \label{tab:ablation}
      \footnotesize
      \setlength{\aboverulesep}{0pt}
      \setlength{\belowrulesep}{0pt}
\vspace{-3mm}     
\begin{tabular}{l|c|c}
    \toprule
    Method  & \thead{AUC of \\ ADD-S} & \thead{AUC of \\ ADD(-S)} \\
    \midrule
    MOTPose &  \textbf{82.0} & \textbf{77.1} \\
    MOTPose without temporal fusion & 80.8 & 74.2 \\
    MOTPose without TEFM & 81.1 & 74.9 \\
    MOTPose without TOFM & 81.4 & 75.3 \\
    MOTPose without SynPick-Ext & 76.4 & 69.2 \\  
    \midrule
    MOTPose\,[$T$=4] & 80.9 & 76.4 \\
    MOTPose\,[$T$=8] & 82.0 &  \textbf{77.1} \\
    MOTPose\,[$T$=12] & \textbf{82.2} & 76.7 \\

    \bottomrule
  \end{tabular}
\end{table}


\subsection{Limitations}
\label{sec:limitations}
Our formulation of multi-object pose estimation as a set prediction problem limits the datasets available for training our model.
Compared to 2D annotations, 6D pose annotations are significantly harder to obtain.
Thus, many of the standard datasets for evaluating object pose estimation like Linemod-Occluded~\citep{Linemodoccluded} and Linemod~\cite{Hinterstoier2012ModelBT} provide pose annotations only for a partial number of objects per scene in the training dataset.
While this is not a limitation for multi-stage methods that process the cropped version of the images for estimating the pose of target objects, our method needs 6D pose annotation for all objects in the scene, which can be prohibitively expensive to acquire in some scenarios.

\section{Conclusion}
We presented MOTPose, a multi-object pose estimation model for RGB video sequences.
Employing the cross-attention-based TEFM and TOFM modules, the MOTPose model fuses object embeddings and object-specific outputs over multiple time steps, respectively.
Aided by the temporal information, our model performs significantly better than the single-frame RGB model while being lighter and significantly faster than other pose tracking methods.

\section{Acknowledgment}
This work has been funded by the German Ministry of
Education and Research (BMBF), grant no. 01IS21080, project
``Learn2Grasp: Learning Human-like Interactive Grasping
based on Visual and Haptic Feedback''.



\printbibliography

@inproceedings{hodavn2020bop,
  title={{BOP} challenge 2020 on {6D} object localization},
  author={Hoda{\v{n}}, Tom{\'a}{\v{s}} and Sundermeyer, Martin and Drost, Bertram and Labb{\'e}, Yann and Brachmann, Eric and Michel, Frank and Rother, Carsten and Matas, Ji{\v{r}}{\'\i}},
  booktitle={European Conference on Computer Vision (ECCV)},
  pages={577--594},
  year={2020},
}

@inproceedings{Liu:EfficientViT:CVPR2023,
  title={{EfficientViT}: Memory Efficient Vision Transformer With Cascaded Group Attention},
  author={Xinyu Liu and Houwen Peng and Ningxin Zheng and Yuqing Yang and Han Hu and Yixuan Yuan},
  booktitle={IEEE/CVF Conference on Computer Vision and Pattern Recognition (CVPR)},
  pages={14420--14430},
  year={2023},
}

@inproceedings{Li:MaskDINO:CVPR2023,
  author       = {Feng Li and
                  Hao Zhang and
                  Huaizhe Xu and
                  Shilong Liu and
                  Lei Zhang and
                  Lionel M. Ni and
                  Heung{-}Yeung Shum},
  title        = {Mask {DINO:} {Towards} a Unified Transformer-based Framework for Object
                  Detection and Segmentation},
  booktitle    = {{IEEE/CVF} Conference on Computer Vision and Pattern Recognition (CVPR)},
  pages        = {3041--3050},
  year         = {2023},
}

@inproceedings{Capellen2020,
  author    = {Catherine Capellen and Max Schwarz and Sven Behnke},
  title     = {{ConvPoseCNN}: {D}ense Convolutional {6D} Object Pose Estimation},
  booktitle = {15th International Conference on Computer Vision Theory and Applications (VISAPP)},
  year      = {2020}
}

@inproceedings{hu2020single,
  title={Single-stage {6D} object pose estimation},
  author={Hu, Yinlin and Fua, Pascal and Wang, Wei and Salzmann, Mathieu},
  booktitle={IEEE/CVF Conference on Computer Vision and Pattern Recognition (CVPR)},
  pages={2930--2939},
  year={2020}
}

@inproceedings{arash2021gcpr,
title= {{T6D-Direct}: Transformers for Multi-Object {6D} Object Pose Estimation},
author={Arash Amini and Arul Selvam Periyasamy and Sven Behnke},
booktitle={DAGM German Conference on Pattern Recognition (GCPR)},
year={2021}}

@inproceedings{labbe2020,
title= {{CosyPose}: Consistent multi-view multi-object {6D} pose estimation},
author={Y. {Labbe} and J. {Carpentier} and M. {Aubry} and J. {Sivic}},
booktitle={European Conference on Computer Vision (ECCV)},
year={2020}}

@inproceedings{li2018deepim,
  title={{DeepIM}: Deep iterative matching for {6D} pose estimation},
  author={Li, Yi and Wang, Gu and Ji, Xiangyang and Xiang, Yu and Fox, Dieter},
  booktitle={European Conference on Computer Vision (ECCV)},
  pages={683--698},
  year={2018}
}

@inproceedings{tekin2018real,
  title={Real-time seamless single shot {6D} object pose prediction},
  author={Tekin, Bugra and Sinha, Sudipta N and Fua, Pascal},
  booktitle={IEEE/CVF Conference on Computer Vision and Pattern Recognition (CVPR)},
  year={2018}
}

@inproceedings{peng2019pvnet,
  title={{PVNet}: Pixel-wise voting network for {6DOF} pose estimation},
  author={Peng, Sida and Liu, Yuan and Huang, Qixing and Zhou, Xiaowei and Bao, Hujun},
  booktitle={IEEE/CVF Conference on Computer Vision and Pattern Recognition (CVPR)},
  pages={4561--4570},
  year={2019}
}

@inproceedings{rad2017bb8,
  title={{BB8}: A scalable, accurate, robust to partial occlusion method for predicting the {3D} poses of challenging objects without using depth},
  author={Rad, Mahdi and Lepetit, Vincent},
  booktitle={IEEE International Conference on Computer Vision (ICCV)},
  pages={3828--3836},
  year={2017}
}

@inproceedings{periyasamy2019refining,
  title={Refining {6D} object pose predictions using abstract render-and-compare},
  author={Periyasamy, Arul Selvam and Schwarz, Max and Behnke, Sven},
  booktitle={IEEE-RAS International Conference on Humanoid Robots (Humanoids)},
  pages={739--746},
  year={2019}
}

@inproceedings{xiang2017posecnn,
    author = {Xiang, Yu and Schmidt, Tanner and Narayanan, Venkatraman and Fox, Dieter},
    title = {{PoseCNN: A} Convolutional Neural Network for {6D} Object Pose Estimation in Cluttered Scenes},
    booktitle   = {Robotics: Science and Systems (RSS)},
    Year = {2018}
}

@inproceedings{manhardt2018deep,
  title={Deep model-based {6D} pose refinement in {RGB} },
  author={Manhardt, Fabian and Kehl, Wadim and Navab, Nassir and Tombari, Federico},
  booktitle={European Conference on Computer Vision (ECCV)},
  pages={800--815},
  year={2018}
}

@inproceedings{hu2019segmentation,
  title={Segmentation-driven {6D} object pose estimation},
  author={Hu, Yinlin and Hugonot, Joachim and Fua, Pascal and Salzmann, Mathieu},
  booktitle={IEEE/CVF Conference on Computer Vision and Pattern Recognition (CVPR)},
  pages={3385--3394},
  year={2019}
}

@inproceedings{Wang_2021_GDRN,
    title = {{GDR-Net}: Geometry-Guided Direct Regression Network for Monocular {6D} Object Pose Estimation},
    author = {Wang, Gu and Manhardt, Fabian and Tombari, Federico and Ji, Xiangyang},
    booktitle = {IEEE/CVF Conference on Computer Vision and Pattern Recognition (CVPR)},
    year = {2021}
}

@inproceedings{periyasamy2018pose,
  author={Periyasamy, Arul Selvam and Schwarz, Max and Behnke, Sven},
  booktitle={IEEE/RSJ International Conference on Intelligent Robots and Systems (IROS)}, 
  title={Robust {6D} Object Pose Estimation in Cluttered Scenes Using Semantic Segmentation and Pose Regression Networks}, 
  year={2018},
  doi={10.1109/IROS.2018.8594406}
  }

@article{kuhn1955hungarian,
  title={The {Hungarian} method for the assignment problem},
  author={Kuhn, Harold W},
  journal={Naval Research Logistics Quarterly},
  volume={2},
  number={1-2},
  pages={83--97},
  year={1955},
  publisher={Wiley Online Library}
}

@inproceedings{schwarz2018fast,
  title={Fast object learning and dual-arm coordination for cluttered stowing, picking, and packing},
  author={Schwarz, Max and Lenz, Christian and Garc{\'\i}a, Germ{\'a}n Mart{\'\i}n and Koo, Seongyong and Periyasamy, Arul Selvam and Schreiber, Michael and Behnke, Sven},
  booktitle={IEEE International Conference on Robotics and Automation (ICRA)},
  pages={3347--3354},
  year={2018},
}

@inproceedings{schwarz2017nimbro,
  title={{NimbRo} picking: Versatile part handling for warehouse automation},
  author={Schwarz, Max and Milan, Anton and Lenz, Christian and Munoz, Aura and Periyasamy, Arul Selvam and Schreiber, Michael and Sch{\"u}ller, Sebastian and Behnke, Sven},
  booktitle={IEEE International Conference on Robotics and Automation (ICRA)},
  pages={3032--3039},
  year={2017},
}

@inproceedings{carion2020end,
  title={End-to-end object detection with transformers},
  author={Carion, Nicolas and Massa, Francisco and Synnaeve, Gabriel and Usunier, Nicolas and Kirillov, Alexander and Zagoruyko, Sergey},
  booktitle={European Conference on Computer Vision (ECCV)},
  pages={213--229},
  year={2020}
}

@article{zhou2020tracking,
  title={Tracking Objects as Points},
  author={Zhou, Xingyi and Koltun, Vladlen and Kr{\"a}henb{\"u}hl, Philipp},
  journal={15th European Conference on Computer Vision (ECCV)},
  year={2020}
}

@InProceedings{tracktor_2019_ICCV,
author = {Bergmann, Philipp and Meinhardt, Tim and Leal{-}Taix{\'{e}}, Laura},
title = {Tracking Without Bells and Whistles},
booktitle = {IEEE International Conference on Computer Vision (ICCV)},
month = {October},
year = {2019}}

@article{Xu2021TransCenterTW,
  title={{TransCenter: Transformers} With Dense Representations for Multiple-Object Tracking},
  author={Yihong Xu and Yutong Ban and Guillaume Delorme and Chuang Gan and Daniela Rus and Xavier Alameda-Pineda},
  journal={IEEE Transactions on Pattern Analysis and Machine Intelligence (TPAMI)},
  year={2021}
}

@article{sun2020transtrack,
  title={{TransTrack}: Multiple-object tracking with transformer},
  author={Sun, Peize and Jiang, Yi and Zhang, Rufeng and Xie, Enze and Cao, Jinkun and Hu, Xinting and Kong, Tao and Yuan, Zehuan and Wang, Changhu and Luo, Ping},
  journal={ arXiv:2012.15460},
  year={2020}
}

@inproceedings{rezatofighi2019generalized,
  title={Generalized intersection over union: A metric and a loss for bounding box regression},
  author={Rezatofighi, Hamid and Tsoi, Nathan and Gwak, JunYoung and Sadeghian, Amir and Reid, Ian and Savarese, Silvio},
  booktitle={IEEE/CVF Conference on Computer Vision and Pattern Recognition (CVPR)},
  pages={658--666},
  year={2019}
}

@InProceedings{Li_2021_CVPR,
author    = {Li, Shichao and Yan, Zengqiang and Li, Hongyang and Cheng, Kwang-Ting},
title     = {Exploring intermediate representation for monocular vehicle pose estimation},
booktitle = {IEEE/CVF Conference on Computer Vision and Pattern Recognition (CVPR)},
year      = {2021},
pages     = {1873-1883}
}

@inproceedings{yolopose2022,
  title={{YOLOPose:} {Transformer-based} multi-object {6D} pose estimation using keypoint regression},
  author={Amini, Arash and Selvam Periyasamy, Arul and Behnke, Sven},
  booktitle={17th International Conference on Intelligent Autonomous Systems (IAS)},
  pages={392--406},
  year={2022},
}

@inproceedings{poserbpf2019,
author    = {Xinke Deng and Arsalan Mousavian and Yu Xiang and Fei Xia and Timothy Bretl and Dieter Fox},
title     = {{PoseRBPF: A} Rao-Blackwellized Particle Filter for {6D} Object Pose Tracking},
booktitle = {Robotics: Science and Systems (RSS)},
year      = {2019}
}

@article{wen2020se,
   title={{se(3)-TrackNet:} Data-driven {6D} Pose Tracking by Calibrating Image Residuals in Synthetic Domains},
   journal={IEEE/RSJ International Conference on Intelligent Robots and Systems (IROS)},
   author={Wen, Bowen and Mitash, Chaitanya and Ren, Baozhang and Bekris, Kostas E.},
   year={2020}}

@inproceedings{trackformer2022,
  title={{TrackFormer}: {Multi-object} tracking with transformers},
  author={Meinhardt, Tim and Kirillov, Alexander and Leal-Taixe, Laura and Feichtenhofer, Christoph},
  booktitle={IEEE/CVF Conference on Computer Vision and Pattern Recognition (CVPR)},
  year={2022}
}

@inproceedings{motr2022,
  title={{MOTR:} {E}nd-to-end multiple-object tracking with transformer},
  author={Zeng, Fangao and Dong, Bin and Zhang, Yuang and Wang, Tiancai and Zhang, Xiangyu and Wei, Yichen},
  booktitle={17th European Conference on Computer Vision (ECCV)},
  year={2022}
}

@article{han2022survey,
  title={A survey on vision transformer},
  author={Han, Kai and Wang, Yunhe and Chen, Hanting and Chen, Xinghao and Guo, Jianyuan and Liu, Zhenhua and Tang, Yehui and Xiao, An and Xu, Chunjing and Xu, Yixing and others},
  journal={IEEE Transactions on Pattern Analysis and Machine Intelligence (TPAMI)},
  volume={45},
  number={1},
  pages={87--110},
  year={2022},
}

@article{khan2022survey,
author = {Khan, Salman and Naseer, Muzammal and Hayat, Munawar and Zamir, Syed Waqas and Khan, Fahad Shahbaz and Shah, Mubarak},
title = {Transformers in Vision: A Survey},
year = {2022},
number       = {10s},
pages        = {200:1--200:41},
issue_date = {January 2022},
publisher = {Association for Computing Machinery},
address = {New York, NY, USA},
volume = {54},
journal = {ACM Computing Survey},
month = {sep},
articleno = {200},
numpages = {41},
}

@inproceedings{wen2023transformers,
  title={Transformers in time series: A survey},
  author={Wen, Qingsong and Zhou, Tian and Zhang, Chaoli and Chen, Weiqi and Ma, Ziqing and Yan, Junchi and Sun, Liang},
  booktitle={32nd International Joint Conference on Artificial Intelligence (IJCAI)},
  year={2023}
}

@article{Hinterstoier2012GradientRM,
  title={Gradient Response Maps for Real-Time Detection of Textureless Objects},
  author={Stefan Hinterstoi{\ss}er and Cedric Cagniart and Slobodan Ilic and Peter F. Sturm and Nassir Navab and Pascal V. Fua and Vincent Lepetit},
  journal={IEEE Transactions on Pattern Analysis and Machine Intelligence (TPAMI)},
  year={2012},
  volume={34},
  pages={876-888}
}

@article{Rothganger20063DOM,
  title={{3D} Object Modeling and Recognition Using Local Affine-Invariant Image Descriptors and Multi-View Spatial Constraints},
  author={Fred Rothganger and Svetlana Lazebnik and Cordelia Schmid and Jean Ponce},
  journal={International Journal of Computer Vision (IJCV)},
  year={2006},
  volume={66},
  pages={231-259}
}

@article{Pavlakos20176DoFOP,
  title={{6-DoF} object pose from semantic keypoints},
  author={Georgios Pavlakos and Xiaowei Zhou and Aaron Chan and Konstantinos G. Derpanis and Kostas Daniilidis},
  journal={IEEE International Conference on Robotics and Automation (ICRA)},
  year={2017},
  pages={2011-2018}
}

@article{Tulsiani2014ViewpointsAK,
  title={Viewpoints and keypoints},
  author={Shubham Tulsiani and Jitendra Malik},
  journal={IEEE/CVF Conference on Computer Vision and Pattern Recognition (CVPR)},
  year={2014},
  pages={1510-1519}
}

@article{Azad20116DoFMT,
  title={{6-DoF} model-based tracking of arbitrarily shaped {3D} objects},
  author={Pedram Azad and David M{\"u}nch and Tamim Asfour and R{\"u}diger Dillmann},
  journal={IEEE International Conference on Robotics and Automation (ICRA)},
  year={2011},
  pages={5204-5209}
}

@article{Pauwels2013RealTimeMR,
  title={Real-Time Model-Based Rigid Object Pose Estimation and Tracking Combining Dense and Sparse Visual Cues},
  author={Karl Pauwels and Leonardo Rubio and Javier D{\'i}az and Eduardo Ros},
  journal={IEEE/CVF Conference on Computer Vision and Pattern Recognition (CVPR)},
  year={2013},
  pages={2347-2354}
}

@InProceedings{yu2014tracking,
author="Xiang, Yu
and Song, Changkyu
and Mottaghi, Roozbeh
and Savarese, Silvio",
title="Monocular Multiview Object Tracking with {3D} Aspect Parts",
booktitle="European Conference on Computer Vision (ECCV)",
year="2014",
pages="220--235"
}

@data{Linemodoccluded,
author = {Brachmann, Eric},
publisher = {heiDATA},
title = {{6D Object Pose Estimation using 3D Object Coordinates [Data]}},
year = {2020},
version = {V1},
doi = {10.11588/data/V4MUMX},
url = {https://doi.org/10.11588/data/V4MUMX}
}

@inproceedings{Hinterstoier2012ModelBT,
  title={Model-based training, detection and pose estimation of texture-less {3D} objects in heavily cluttered scenes},
  author={Hinterstoisser, Stefan and Lepetit, Vincent and Ilic, Slobodan and Holzer, Stefan and Bradski, Gary and Konolige, Kurt and Navab, Nassir},
  booktitle={Asian Conference on Computer Vision (ACCV)},
  pages={548--562},
  year={2013},
}

@inproceedings{periyasamy2021synpick,
  title={{SynPick: A} dataset for dynamic bin picking scene understanding},
  author={Periyasamy, Arul Selvam and Schwarz, Max and Behnke, Sven},
  booktitle={IEEE International Conference on Automation Science and Engineering (CASE)},
  pages={488--493},
  year={2021}
}

@inproceedings{jantos2023poet,
  title={{PoET}: {Pose} Estimation Transformer for Single-View, Multi-Object {6D} Pose Estimation},
  author={Jantos, Thomas Georg and Hamdad, Mohamed Amin and Granig, Wolfgang and Weiss, Stephan and Steinbrener, Jan},
  booktitle={Conference on Robot Learning (CoRL)},
  pages={1060--1070},
  year={2023},
  organization={PMLR}
}

@inproceedings{pyrapose2021,
  author       = {Stefan Thalhammer and
                  Markus Leitner and
                  Timothy Patten and
                  Markus Vincze},
  title        = {{PyraPose}: Feature Pyramids for Fast and Accurate Object Pose Estimation
                  under Domain Shift},
  booktitle    = {IEEE International Conference on Robotics and Automation (ICRA)},
  pages        = {13909--13915},
  year         = {2021},
}

@inproceedings{castro2023crt,
  title={{CRT-6D: Fast 6D} object pose estimation with cascaded refinement transformers},
  author={Castro, Pedro and Kim, Tae-Kyun},
  booktitle={IEEE/CVF Winter Conference on Applications of Computer Vision (WACV)},
  pages={5746--5755},
  year={2023}
}

@inproceedings{hai2023shape,
  title={Shape-Constraint Recurrent Flow for {6D} Object Pose Estimation},
  author={Hai, Yang and Song, Rui and Li, Jiaojiao and Hu, Yinlin},
  booktitle={Proceedings of the IEEE/CVF Conference on Computer Vision and Pattern Recognition (CVPR)},
  pages={4831--4840},
  year={2023}
}

@inproceedings{hu2022perspective,
  title={Perspective flow aggregation for data-limited {6D} object pose estimation},
  author={Hu, Yinlin and Fua, Pascal and Salzmann, Mathieu},
  booktitle={European Conference on Computer Vision (ECCV)},
  pages={89--106},
  year={2022},
  organization={Springer}
}

@inproceedings{di2021so,
  title={{SO-Pose: E}xploiting self-occlusion for direct {6D} pose estimation},
  author={Di, Yan and Manhardt, Fabian and Wang, Gu and Ji, Xiangyang and Navab, Nassir and Tombari, Federico},
  booktitle={IEEE/CVF International Conference on Computer Vision (ICCV)},
  pages={12396--12405},
  year={2021}
}

@article{periyasamy2023yolopose,
  title={{YOLOPose V2: U}nderstanding and improving transformer-based {6D}‚ pose estimation},
  author={Periyasamy, Arul Selvam and Amini, Arash and Tsaturyan, Vladimir and Behnke, Sven},
  journal={Robotics and Autonomous Systems},
  volume={168},
  pages={104490},
  year={2023},
  publisher={Elsevier}
}

@article{periyasamy2023efficient,
  title={Efficient Multi-Object Pose Estimation using Multi-Resolution Deformable Attention and Query Aggregation},
  author={Periyasamy, Arul Selvam and Tsaturyan, Vladimir and Behnke, Sven},
  journal={IEEE International Conference on Robotic Computing (IRC)},
  year={2023}
}

\end{document}